\documentclass[10pt,twocolumn,letterpaper]{article}

\usepackage{cvpr}      

\usepackage{graphicx}
\usepackage{amsmath}
\usepackage{amssymb}
\usepackage{booktabs}
\usepackage{mmstyles}
\usepackage{bbm}
\usepackage[table]{xcolor}
\usepackage{colortbl}
\usepackage{multirow}
\definecolor{myy}{RGB}{126,95,0}
\definecolor{mygray}{gray}{.9}
\definecolor{bblue}{RGB}{30,80,120}
\definecolor{mygray1}{gray}{.7}
\definecolor{ggray}{RGB}{127,127,127}
\definecolor{mygreen}{RGB}{93,174,86}

\usepackage[pagebackref,breaklinks,colorlinks]{hyperref}

\usepackage[capitalize]{cleveref}
\crefname{section}{Sec.}{Secs.}
\Crefname{section}{Section}{Sections}
\Crefname{table}{Table}{Tables}
\crefname{table}{Tab.}{Tabs.}

\def\cvprPaperID{6425} 
\def\confName{CVPR}
\def\confYear{2022}

\begin{document}

\title{Contrastive learning of Class-agnostic Activation Map for Weakly Supervised Object Localization and Semantic Segmentation}

\author{Jinheng Xie, Jianfeng Xiang, Junliang Chen, Xianxu Hou,  Xiaodong Zhao, Linlin Shen\thanks{Corresponding Author}\\
	School of Computer Science \& Software Engineering, Shenzhen University, China\\
	Dept. of Computer Science, Wenzhou-Kean University, Wenzhou, China \\
	Shenzhen Institute of Artificial Intelligence of Robotics of Society, Shenzhen, China \\
	Guangdong Key Laboratory of Intelligent Information Processing, Shenzhen University, China\\
	{\tt\small xiejinheng2020@email.szu.edu.cn, hxianxu@gmail.com,  llshen@szu.edu.cn}
}
\maketitle

\begin{abstract}

    While class activation map (CAM) generated by image classification network has been widely used for weakly supervised object localization (WSOL) and semantic segmentation (WSSS), such classifiers usually focus on discriminative object regions. In this paper, we propose Contrastive learning for Class-agnostic Activation Map (C$^2$AM) generation only using unlabeled image data, without the involvement of image-level supervision. The core idea comes from the observation that i) semantic information of foreground objects usually differs from their backgrounds; ii) foreground objects with similar appearance or background with similar color/texture have similar representations in the feature space. We form the positive and negative pairs based on the above relations and force the network to disentangle foreground and background with a class-agnostic activation map using a novel contrastive loss. As the network is guided to discriminate cross-image foreground-background, the class-agnostic activation maps learned by our approach generate more complete object regions. We successfully extracted from C$^2$AM class-agnostic object bounding boxes for object localization and background cues to refine CAM generated by classification network for semantic segmentation. Extensive experiments on CUB-200-2011, ImageNet-1K, and PASCAL VOC2012 datasets show that both WSOL and WSSS can benefit from the proposed C$^2$AM. Code will be available at \href{https://github.com/CVI-SZU/CCAM}{https://github.com/CVI-SZU/CCAM}.
\end{abstract}

\section{Introduction}
\label{sec:intro}
\begin{figure}[t]
	\begin{center}
		\includegraphics[width=\linewidth]{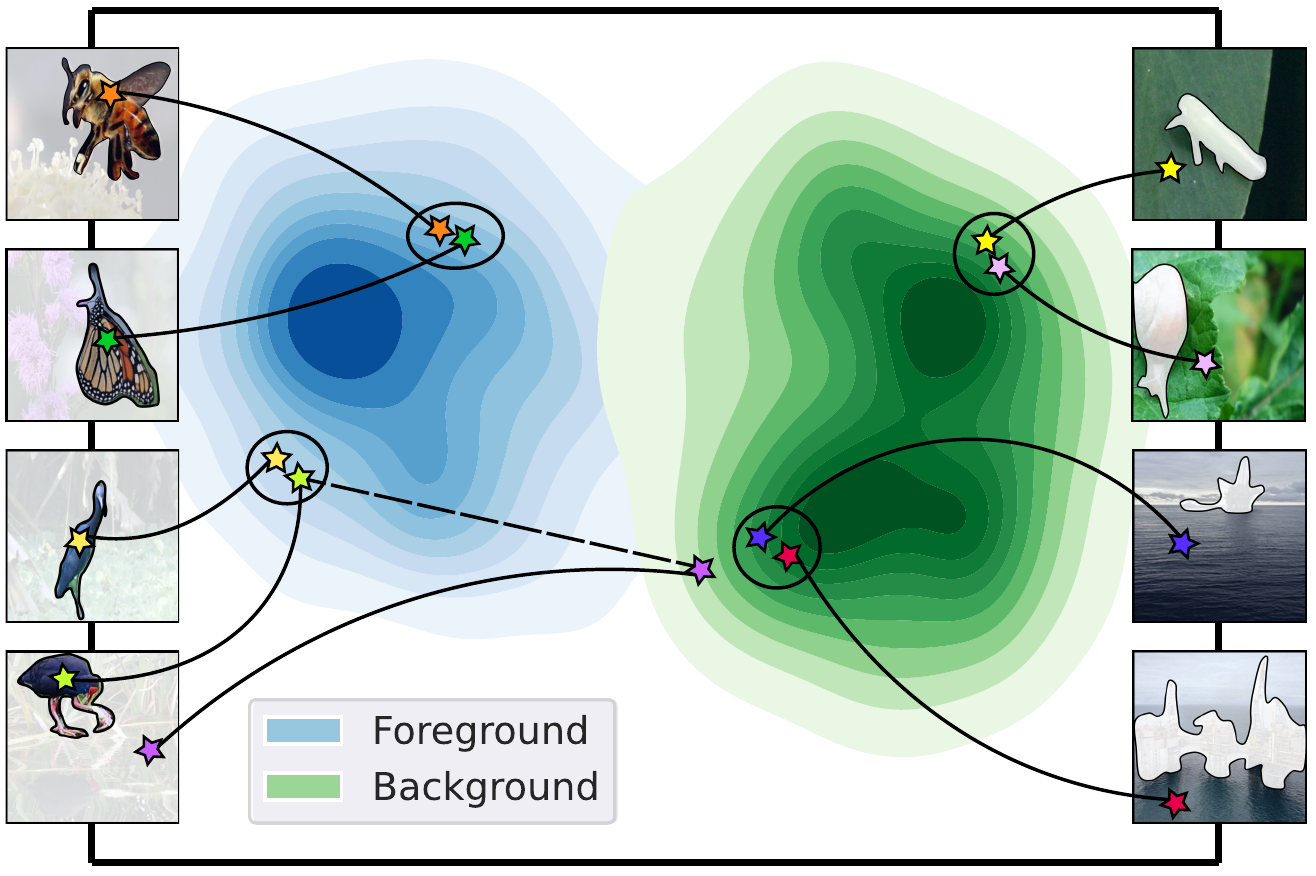}
	\end{center}
	\vspace{-18pt}
	\caption{Feature manifold of foreground objects (blue) and backgrounds (green). As semantic information of foreground objects differs from that of backgrounds, the distribution of the representation of foreground objects (blue) is far away from backgrounds (green). Foreground objects with similar appearance or backgrounds with similar color/texture also have similar representations in the feature space. Based on these observations, positive and negative pairs can be formed for contrastive learning. t-SNE~\cite{tsne} is used to reduce the dimensionality of features.}
	\label{fig:insight}
	\vspace{-10pt}
\end{figure}
Massive image data and manual annotations are usually required to train deep neural networks for many vision tasks, e.g., object detection and semantic segmentation. However, it is time-consuming and labor-intensive to obtain bounding box or pixel-level annotations. In recent years, \textit{weaker supervision}, e.g., image-level label, has been introduced in weakly supervised object localization (WSOL) and semantic segmentation (WSSS), which aims to achieve localization or segmentation with \ only image-level supervision, i.e., without bounding box or pixel-level annotations. Most previous WSOL and WSSS methods rely on class activation map (CAM) to estimate location of the target object. With image-level supervision, the classifier tries to find discriminative regions of target objects. Thus, though image-level labels enable CAM to indicate the correct location of target objects, they also limit the focus of CAM on sparse and discriminative object regions. As a result, it could be very difficult to precisely estimate complete object regions by directly applying CAM to WSOL and WSSS.

\begin{figure}[t]
	\begin{center}
		\includegraphics[width=0.95\linewidth]{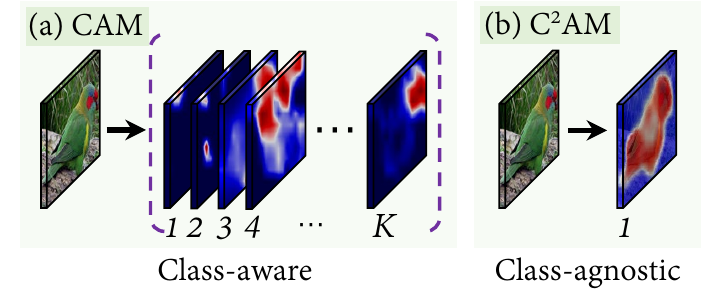}
	\end{center}
	\vspace{-18pt}
	\caption{Difference between (a) the class activation map (CAM) and (b) class-agnostic activation map. CAM consists of $K$ (number of class) activation maps and C$^2$AM only predicts one class-agnostic activation map for an image, which directly indicates the foreground and background regions. Best viewed in color.}
	\label{fig:cam_ccam_compar}
	\vspace{-14pt}
\end{figure}

Different methods have been introduced to mitigate the above problem of CAM. However, most of them are trained with image-level supervision, which would potentially affect the completeness of CAM. By contrast, this paper proposes to generate class-agnostic activation maps based on a novel \textit{cross-image foreground-background contrastive learning}, without the requirement of image-level supervision. In contrast with $K$ class activation maps (CAM) (including one target activation map), C$^2$AM only predicts one class-agnostic activation map to indicate the foreground and background regions in an image. The difference between CAM and C$^2$AM is illustrated in Figure \ref{fig:cam_ccam_compar}. Compared with CAM, the class-agnostic activation maps learned from \textit{cross-image foreground-background contrast}, without the involvement of any image-level supervision, produce more reliable foreground regions. 

\begin{figure*}[t]
	\begin{center}
		\includegraphics[width=\linewidth]{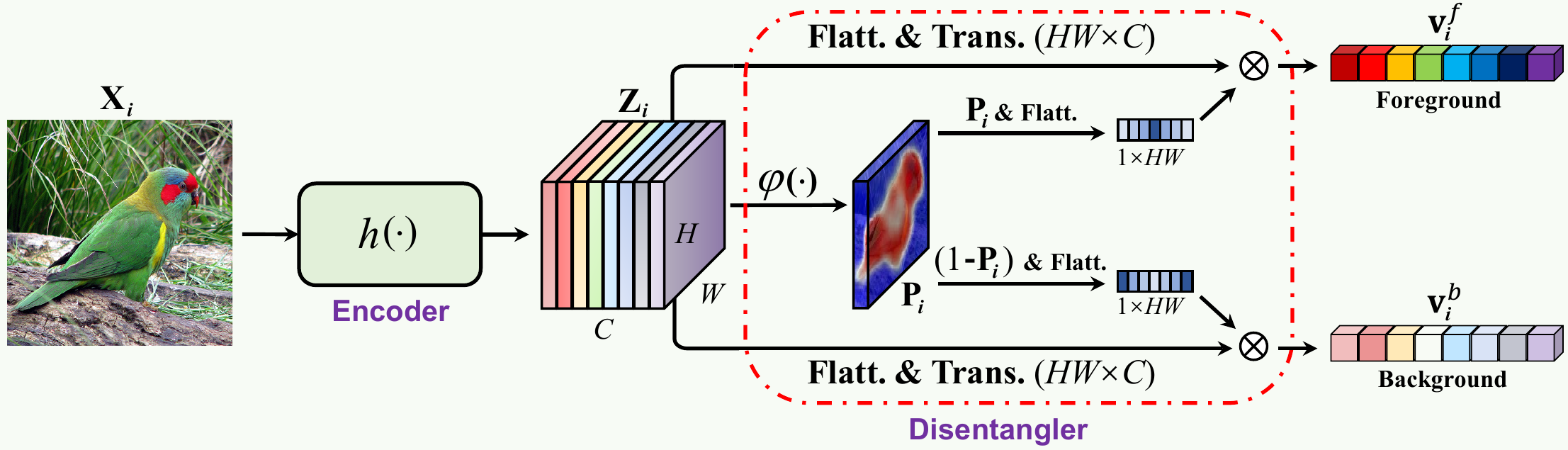}
	\end{center}
	\vspace{-18pt}
	\caption{The overall network architecture of the proposed method. The encoder network $h(\cdot)$ maps image $\mX_i$ to the feature map $\mZ_i$. In disentangler, the activation head $\varphi(\cdot)$ produces the class-agnostic activation map $\mP_i$. Suppose $\mP_i$ activates the foreground regions and the background activation map can be derived as $(1-\mP_i)$. Based on the foreground and background activation maps, $\mZ_i$ can be disentangled into the foreground and background feature representations, i.e., $\vv^f_i$ and $\vv^b_i$. In evaluation, only the trained $h(\cdot)$ and $\varphi(\cdot)$ are used for generating the class-agnostic activation map $\mP_i$. \textbf{Flatt.}: matrix flattening; \textbf{Trans.}: matrix transpose; $\otimes$: matrix multiplication. }
	\label{fig:network}
	\vspace{-10pt}

\end{figure*}

\begin{figure*}[t]
	\begin{center}
		\includegraphics[width=0.97\linewidth]{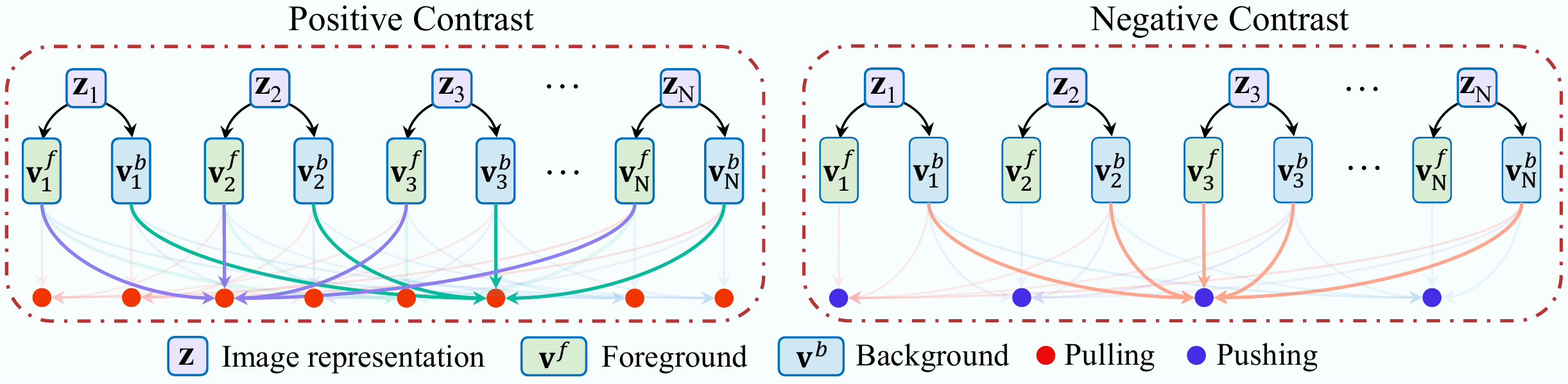}
	\end{center}
	\vspace{-18pt}
	\caption{Illustration of cross-image foreground-background contrastive learning. Each image representation, i.e., $\mZ_i$, is disentangled into the foreground and background representations, i.e., $\vv_i^f$ and $\vv_i^b$. Two foreground or background representations are coupled into one positive pair, while a negative pair is formed with one foreground and one background representation. Contrastive learning is applied to pull close the representations from the positive pair and push apart the representations from the negative pair. Best viewed in color.}
	\label{fig:losses}
	\vspace{-10pt}
\end{figure*}

As shown in Figure \ref{fig:insight}, the semantic information of the foreground object differs from its background, which can be presented as the large distance between foreground and background representations in the feature space. This can be further extended to the cross-image cases where the semantic information of the foreground object from one image should also be largely different from that of background in other images. By contrast, the distance among foregrounds with similar appearance or backgrounds with similar color/texture shall be small. Based on the above observations, we propose the cross-image foreground-background contrastive loss to force the network to disentangle the foreground object and background in an image with a class-agnostic activation map. Specifically, as shown in Figure \ref{fig:network} and \ref{fig:losses}, the network first generates a class-agnostic activation map using an activation head, such that the image representation can be disentangled into the foreground and background representations, respectively. Then, the foreground and background representations form the negative pairs and the foreground-foreground or background-background representations form the positive pairs. As only foregrounds with similar appearance or backgrounds with similar color/texture have similar representations in the feature space, pairs of foreground-foreground or background-background sharing less similar semantics might affect the learning of network. To mitigate this problem, we design a feature similarity based rank weighting to automatically reduce the influence of those dissimilar positive pairs. As the activation head is randomly initialized, the initial class-agnostic activation maps are random as well at beginning. When contrastive loss is applied to pull close and push apart the representations of positive and negative pairs, the class-agnostic activation map gradually separates the regions of foreground object and background in the image. 

Extensive experiments conducted on the tasks of WSOL and WSSS show that the proposed C$^2$AM can replace or refine CAM for better performance. Specifically in WSOL, we follow \cite{psol} to divide WSOL into two tasks: class-agnostic object localization and object classification, and thus the class-agnostic activation maps can be used to extract class-agnostic object bounding boxes for localization. Background cues can also be extracted from the class-agnostic activation maps to refine the initial CAM such that false activation of background can be efficiently reduced to generate more reliable object regions. The refined CAM can thus greatly enhance the following segmentation performance in WSSS.

Collectively, the main contributions of this paper can be summarized as:
\begin{itemize}
	\item We propose cross-image foreground-background contrastive learning to generate class-agnostic activation maps with unlabeled image data, which doesn't involve any image-level supervision and produces much more reliable foreground object regions than CAM.
	
	\item The class-agnostic activation maps can be used to extract class-agnostic bounding boxes for accurate object localization and serve as background cues to improve the quality of initial CAM. 
	
	\item Extensive experiments show that both WSOL and WSSS tasks can benefit from our method. We provide an alternative solution to replace or refine CAM for improvements of weakly supervised learning. Besides, it might be further used in a lot of vision tasks to detect foreground regions without manual annotations.  
\end{itemize}

\section{Related Works}
Class activation mapping (CAM)~\cite{zhou2016learning} produces a spatial class activation map that indicates the object regions responsible for the prediction of object category. However, the classifier tends to perform recognition based on the most discriminative regions of the object~\cite{vgg, resnet}, which affects the precision of CAM based localization or segmentation.

\textbf{Weakly supervised object localization (WSOL).} Most of existing WSOL works consider the image-level labels in the training dataset~\cite{cubwah2011caltech,imagenet}.  To solve the above problem, various methods~\cite{hassingh2017hide, adlchoe2019attention, spgzhang2018self, i2c, GCNWSOJ, acol, danet, ornet} have been proposed. Baek \etal~\cite{psynet} introduce class-agnostic activation mapping, which can effectively generate heat map from a self-supervised model. Xie \etal~\cite{ornet} propose a low-level feature based two-stage learning framework to refine coarse activation maps. Differently, Zhang \etal~\cite{psol} divide WSOL into two tasks: 1) class-agnostic object localization and 2) object classification. This allows a network to complete localization or classification separately. They use DDT~\cite{ddt} to extract class-agnostic object bounding boxes as pseudo labels to train a localization model and use state-of-the-art networks for classification. 

\textbf{Weakly supervised semantic segmentation (WSSS).} Most WSSS works follow a three-stage learning process: initial CAM generation, pseudo masks generation, and segmentation model training. Hou \etal~\cite{seenet} propose two self-erasing strategies to focus attention only on the reliable regions and generate complete initial CAM. Chang \etal \cite{sc-cam} propose to mine more object parts by investigating object sub-categories. Jungbeom \etal~\cite{advcam} propose an anti-adversarial manner to discover more regions of the target object in the activation map. Ahn and Kwak \cite{affinitynet} propose pixel-level semantic affinity, which helps to generate fine-grained pseudo masks. Works~\cite{oaa, mcis, liid, eps, edam} additionally use background cues from a fully supervised saliency detector to get better pseudo masks. In our work, the class-agnostic activation maps generated by C$^2$AM are used to extract background cues to refine the initial CAM, which increases the potential of initial CAM for the next stage. 

\textbf{Contrastive learning.} The idea of contrastive learning is to pull close the samples from the positive pair and to push apart the samples from the negative pair~\cite{siamese, contrastivelecun, triplet, moco, simclr}.
Based on object class labels, the positive pair is created with the samples from the same class, while the samples with different class labels form the negative pair~\cite{siamese, contrastivelecun, triplet, supcontrast}. Unsupervised contrastive learning can be divided into two categories, i.e., instance-wise contrastive learning~\cite{instanceloss, isif, moco, simclr} and clustering based contrastive learning~\cite{ca, jul, prototypical}. In instance-wise contrastive learning, two samples augmented from the same instance form the positive pair, while samples augmented from different instances~\cite{moco, simclr} become the negative pair. For clustering based contrastive learning, the clustering algorithm is applied to generate the pseudo label for training samples, and then the supervised contrastive learning is applied~\cite{self-paced}. 

\section{Methodology}

\subsection{Architecture}
The overall network architecture of C$^2$AM is shown in Figure \ref{fig:network}.
{Given a batch of $n$ images $\mX_{1:n}=\{\mX_i\}_{i=1}^{n}$, the encoder $h(\cdot)$ maps $\mX$ to high-level feature maps $\mZ_{1:n}=\{\mZ_i\}_{i=1}^n$, in which $\mZ_i \in \mathbb{R}^{C\times H\times W}$. $C$ and $H\times W$ denote the channel number and spatial dimension, respectively.
The popular network, e.g., ResNet~\cite{resnet} or VGG~\cite{vgg}, is used as the encoder $h(\cdot)$. Supervised or unsupervised pretraining, e.g., moco~\cite{moco} and detco~\cite{detco}, on ImageNet-1K~\cite{imagenet} can be adopted as initialization of $h(\cdot)$. 
Based on the extracted feature maps $\mZ$, the disentangler employs an activation head $\varphi(\cdot)$ to produce the class-agnostic activation maps $\mP_{1:n}=\{\mP_i\}_{i=1}^{n}$, in which $\mP_i\in \mathbb{R}^{1\times H\times W}$. Specifically, $\varphi(\cdot)$ is a 3$\times$3 convolution with a batch normalization layer. Suppose $\mP_i$ activates foreground regions, the background activation map of the $i$-th sample can be formulated as $(1-\mP_i)$.
The foreground and background activation maps can finally disentangle the feature map $\mZ_{1:n}$ into the foreground and background feature representations, i.e., $\vv^f_{1:n}$ and $\vv^b_{1:n}$, respectively. For the $i$-th sample, $\vv^f_i$ and $\vv^b_i$ can be derived as:
\begin{equation}
	\label{eq:fg_bg_features}
	\vv_i^f= \mP_i \otimes \mZ_i^\top, \quad \vv_i^b = (1-\mP_i) \otimes \mZ_i^\top.
\end{equation}
Here, $\mP_i$ and $\mZ_i$ are flattened, i.e., $\mP_i \in \mathbb{R}^{1\times HW}$ and $\mZ_i \in \mathbb{R}^{C\times HW}$. $\vv^f_i\in \mathbb{R}^{1\times C}$ and $\vv^b_i\in \mathbb{R}^{1\times C}$.  $\otimes$ and $\top$ indicate the matrix multiplication and transpose, respectively.

\subsection{Foreground-background Contrast}
Considering the fact that there is no label information in the training process, C$^2$AM is proposed based on the idea of \textit{cross-image foreground-background contrast}, which locates the foreground object regions by only utilizing the semantic information among foreground and background representations. As aforementioned, given an image, its foreground and background representations contain different semantic information and thus should have a large distance in the feature space. This observation is also true for cross-image cases.
The distance between their foreground and background representations should also be large.

Based on these analyses, we propose to apply the contrastive learning to push apart foreground-background representations. Given $n$ samples $\mX_{1:n}$, the disentangler separates them into $n$ foreground and $n$ background representations, i.e., $\vv^f_{1:n}$ and $\vv^b_{1:n}$. The foreground-background representation pair, i.e., $(\vv^f_i,\vv^b_j)$, is treated as the \textit{negative pair}. The negative contrastive loss is designed as:
\begin{equation}
	\label{eq:l_neg}
	\cL_{NEG} = -\frac{1}{n^2 } \sum_{i=1}^{n}\sum_{j=1}^{n}\log (1 - s^{neg}_{i,j}),
\end{equation}
\begin{equation}
	s^{neg}_{i,j} = \text{sim}(\vv^f_{i}, \vv^b_{j}),
\end{equation}
where $s^{neg}_{i,j}$ is the cosine similarity between $\vv^f_{i}$ and $\vv^b_{j}$. $\cL_{NEG}$ considers the contrastive comparisons both within image ($i=j$) and cross-image ($i\neq j$).

\subsection{Foreground-foreground and Background-background Contrast with Rank Weighting}
The foreground and foreground or background and background representations from two different images form the \textit{positive pair}. However, only representations of foregrounds with similar appearance and background with similar color/texture have a smaller distance in the feature space and shall be pulled together. Positive pairs with large distances will affect the learning process, as there is fewer similar semantics in these two foreground objects or backgrounds. To address this issue, we design a feature similarity based rank weighting to automatically reduce the influence of those dissimilar positive pairs. We first calculate the cosine similarity between each candidate positive pair: 
\begin{equation}
	s^{f}_{i,j} =  \text{sim}(\vv^f_{i},\vv^f_{j}), 
	s^{b}_{i,j} =  \text{sim}(\vv^b_{i}, \vv^b_{j}),
\end{equation}
where $s^{f}_{i,j}$ and $s^{b}_{i,j}$ are the cosine similarity calculated from the foreground-foreground pair, i.e., $(\vv^f_i,\vv^f_j)$, and background-background pair, i.e., $(\vv^b_i,\vv^b_j)$, respectively. $s^{f}_{i,j}$ potentially indicates whether the foreground object from $\mX_i$ shares similar semantics with that from $\mX_j$. Given the set of similarity from foreground-foreground representations $s^f=\{s^{f}_{1,2},\cdots, s^{f}_{i,j}, \cdots\}(i \neq j)$ and the set of similarity from background-background representations $s^b=\{s^{b}_{1,2},\cdots, s^{b}_{i,j}, \cdots\}(i \neq j)$, we then calculate a weight based on ranking for each positive pair as follow:
\begin{equation}
	\label{eq:rank_weighting}
	w^{f}_{i,j} =  \exp(-\alpha\cdot \text{rank}( s^{f}_{i,j})), w^{b}_{i,j} =  \exp(-\alpha\cdot\text{rank}( s^{b}_{i,j})),
\end{equation}
where $\alpha$ is a hyper-parameter controlling the smoothness of exponential function. rank($s^f_{i,j}$) and rank($s^b_{i,j}$) are the rank of $s^f_{i,j}$ and $s^b_{i,j}$ in the set of $s^f$ and $s^b$, respectively. The rank weight $w_{i,j}$ ranges from 0 to 1. A large weight is assigned to positive pairs sharing similar semantics (e.g., similar appearance, color or texture) and a small one is assigned to the positive pairs with less similarity. It can reduce the influence of those dissimilar pairs to some extent for better contrastive learning. The final positive contrastive loss is formulated as: 
\begin{equation}
	\label{eq:l_pos}
	\cL_{POS} = \cL_{POS}^f + \cL_{POS}^b,
\end{equation}
\begin{equation}
	\label{eq:l_pos_f}
	\cL_{POS}^f = -\frac{1}{n(n-1)} \sum_{i=1}^{n}\sum_{j=1}^{n} \mathbbm{1}_{[i\neq j]} (w^f_{i,j} \cdot\log( s^{f}_{i,j})),
\end{equation}
\begin{equation}
	\label{eq:l_pos_b}
	\cL_{POS}^b = -\frac{1}{n(n-1)} \sum_{i=1}^{n}\sum_{j=1}^{n} \mathbbm{1}_{[i\neq j]} (w^b_{i,j} \cdot\log( s^{b}_{i,j})), 
\end{equation}
where $\mathbbm{1}_{[i\neq j]} \in \{0,1\}$ is an indicator function and it equals 1 if $i \neq j$. The overall contrastive loss $\cL$ is formulated as the summation of $\cL_{POS}$ and $\cL_{NEG}$:
\begin{equation}
	\label{eq:cfc_loss}
	\cL = \cL_{POS} + \cL_{NEG}.
\end{equation}
When contrastive loss $\mathcal{L}$ is applied to pull close and push apart the representations in positive and negative pairs, the class-agnostic activation map gradually separates the regions of foreground object and background in the image.

\textbf{How to determine foreground regions.} As th same contrastive loss is applied to foreground-foreground and background-background positive pairs, it can not guarantee that foreground or background regions are activated in $\mP_i$. To solve this problem, we set a threshold to binarize the class-agnostic activation maps and detect the largest contour to determine the object regions.

\subsection{Weakly Supervised Object Localization}
We follow the route of PSOL~\cite{psol} to adapt our C$^2$AM to WSOL. Specifically, WSOL is divided into two tasks: class-agnostic object localization and object classification. PSOL uses DDT~\cite{ddt} to generate class-agnostic object bounding boxes on the training set. Given a group of images of the same category, DDT obtains those category-consistent regions to extract class-agnostic object bounding boxes. By contrast, the proposed C$^2$AM directly learns class-agnostic activation maps from the whole dataset without any manual annotations. We set a threshold to binarize the class-agnostic activation maps and then extract class-agnostic object bounding boxes as pseudo labels (we present a fair comparison between DDT and C$^2$AM in the next section). The localization model is trained with these pseudo labels for object bounding box predictions. The popular network, e.g., EfficientNet~\cite{efficientnet}, is adopted for object classification. 

\subsection{Weakly Supervised Semantic Segmentation}
\label{sec:wsss}
We first use CAM-based methods to generate initial CAM for each image and then apply C$^2$AM to refine it. Specifically, we use the background activation maps $(1-\mP)$ as pseudo labels to further train a model to predict the background regions, i.e., background cues, in the image. As shown in Figure \ref{fig:app}, we concatenate the predicted background cues with the initial CAM and perform the argmax process along the channel dimension to refine initial CAM. This helps to reduce the false activation of background and activate more foreground regions in initial CAM. More details of the pseudo training and refinement procedure are included in supplementary materials. We just use a simple way to demonstrate the effectiveness of our C$^2$AM for refinement of initial CAM and leave more works in the future.  
\begin{figure}[t]
	\begin{center}
		\includegraphics[width=0.9\linewidth]{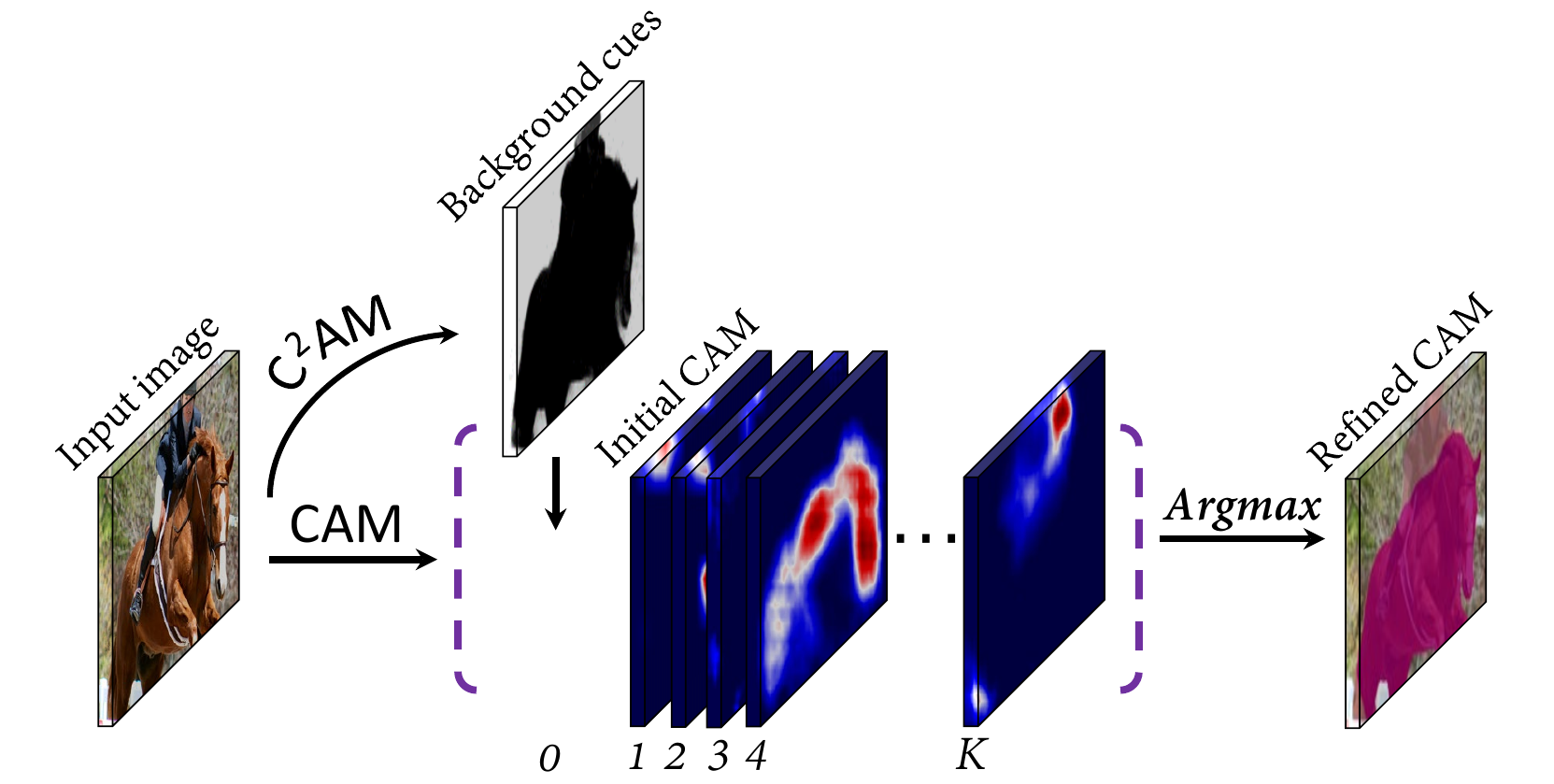}
	\end{center}

	\vspace{-18pt}
	\caption{Refinement of initial CAM using background cues.}
		\label{fig:app}
	\vspace{-16pt}
\end{figure}
\begin{figure*}[ht]
	\begin{center}
		\includegraphics[width=0.95\linewidth]{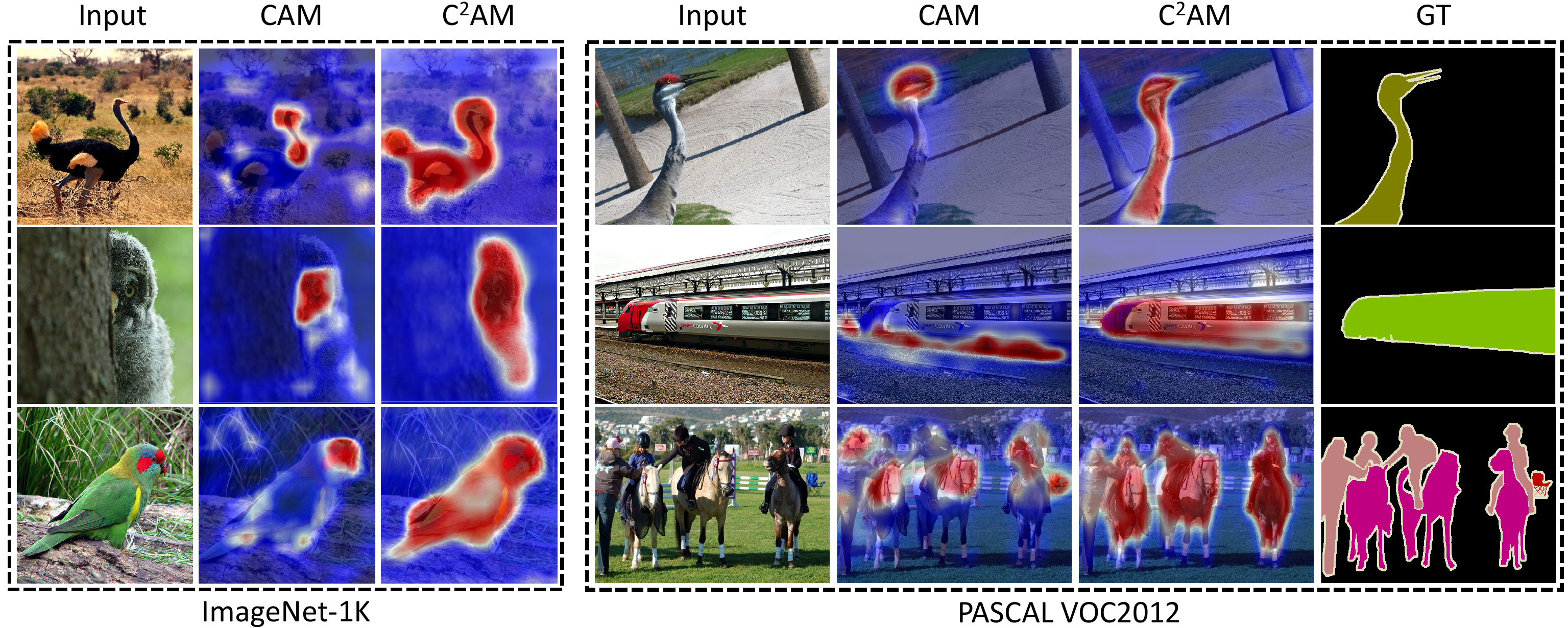}
	\end{center}
	\vspace{-18pt}
	\caption{Visual comparison between CAM and the class-agnostic activation maps generated by C$^2$AM. Best viewed in color.}
	\label{fig:visual_compar}
	\vspace{-5pt}
\end{figure*}
\vspace{-14pt}
\section{Experiments}
\subsection{Experimental Setup}
\textbf{Dataset.} CUB-200-2011~\cite{cubwah2011caltech} is a fine-grained classification dataset, which consists of 200 species of birds with 5,994 images for training and 5,794 images for test. ImageNet-1K~\cite{imagenet} is a large scale visual recognition dataset with 1,000 classes, which contains 1,281,197 training images and 50,000 validation images. PASCAL VOC2012~\cite{pascal-voc-2012} is a popular semantic segmentation dataset with 20 object categories. It consists of 1,464 images for training, 1,449 images for validation, and 1,456 images for test. 

\textbf{Evaluation Metrics.} In WSOL, we follow ~\cite{zhou2016learning} to use Top-1 localization accuracy (Top-1 Loc), Top-5 localization accuracy (Top-5 Loc), and GT-known localization accuracy (GT-known Loc) for evaluation. For GT-known Loc, a localization is correct when the predicted bounding box overlaps over 50\% with one of the ground truth bounding boxes belonging to the same class. For Top-1 Loc, a prediction is correct when the Top-1 classification result and GT-known Loc are both correct. For Top-5 Loc, a prediction is correct when one of the Top-5 classification results and GT-known Loc are both correct. The MaxBoxAccV2 proposed in \cite{ewsol} is also adopted, which averages the performance across different bounding boxes ratios, i.e., 30\%, 50\%, and 70\%. In WSSS, the Intersection over Union (IoU) and mean Intersection over Union (mIoU) is adopted as the evaluation metric. 

The implementation details are provided in the supplementary materials.

\begin{table*}[t]
	\centering
	\caption{Comparison of the performance between the proposed method and the state-of-the-art methods on CUB-200-2011 test set and ImageNet-1K validation set. Loc Back. denotes the localization backbone. Cls Back. denotes the backbone for classification. $^\dag$ and $^{\ddag}$ indicate the $h(\cdot)$ is initialized by the supervised and unsupervised (moco) pretraining, respectively. $^*$: our re-implementation results.}
	\vspace{-5pt}
	\resizebox{\linewidth}{!}{
		\begin{tabular}{l c c c c c c c c}
			\toprule[1.5pt]
			\multirow{2.5}{*}{Method} & \multirow{2.5}{*}{Loc Back.} &  \multirow{2.5}{*}{Cls Back.} & \multicolumn{3}{c}{CUB-200-2011} & \multicolumn{3}{c}{ImageNet-1K}\\
			\cmidrule(lr){4-6}
			\cmidrule(lr){7-9}
			& & & Top-1 Loc & Top-5 Loc  & GT-known Loc & Top-1 Loc & Top-5 Loc  & GT-known Loc\\
			\midrule
			CAM$_\text{ CVPR'16}$~\cite{zhou2016learning} & \multicolumn{2}{c}{VGG-GAP} & 36.13 & - & - & 42.80 & 54.86 & 59.00  \\
			ADL$_\text{ CVPR'19}$~\cite{adlchoe2019attention} & \multicolumn{2}{c}{VGG-GAP} & 52.36 & - & 73.96 & 44.92 & - & - \\
			I$^2$C$_\text{ ECCV'20}$~\cite{i2c} & \multicolumn{2}{c}{InceptionV3} & 65.99 & 68.34  & 72.60 & 53.11 & 64.13 & 68.50 \\
			GC-Net$_\text{ ECCV'20}$~\cite{GCNWSOJ} & \multicolumn{2}{c}{GoogLeNet} & 58.58 & 71.10  & 75.30 & 49.06 & 58.09 & - \\
			SPA$_\text{ CVPR'21}$~\cite{spa} & \multicolumn{2}{c}{VGG16} & 60.27 & 72.5  & 77.29 & 49.56 & 61.32 & 65.05 \\
			FAM$_\text{ ICCV'21}$~\cite{fam} & \multicolumn{2}{c}{VGG16} & 69.26 & -  & 89.26 & 51.96 & - & \textbf{71.73} \\
			ORNet$_\text{ ICCV'21}$~\cite{ornet} & \multicolumn{2}{c}{VGG16} & 67.74 & 80.77  & 86.2 & 52.05 & 63.94 & 68.27 \\
			CAM$_\text{ CVPR'16}$~\cite{zhou2016learning} & \multicolumn{2}{c}{DenseNet161} & 29.81 & 39.85 & - & 39.61 & 50.40 & 52.54 \\
			\midrule
			PSOL$_\text{ CVPR'20}$~\cite{psol} & ResNet50  & ResNet50 & 70.68 & 86.64  & 90.00 & 53.98 & 63.08 & 65.44 \\
			PSOL$_\text{ CVPR'20}$~\cite{psol} & DenseNet161  & EfficientNet-B7 & 80.89$^*$ & 89.97$^*$  & 91.78$^*$ & 58.00 & 65.02 & 66.28 \\
			
			\midrule
			\rowcolor{mygray}
			C$^2$AM(Ours)$^{\dag}$ & ResNet50  & ResNet50 & 76.36  &  89.15  &  93.40 & 54.41 &  64.77 & 67.80 \\
			\rowcolor{mygray}
			C$^2$AM(Ours)$^{\ddag}$ & ResNet50  & ResNet50 & 74.76 & 87.37 & 91.54 & 54.65 & 65.05 & 68.07 \\
			\rowcolor{mygray1}
			C$^2$AM(Ours)$^\dag$ & DenseNet161  & EfficientNet-B7 & \textbf{83.28}  &  \textbf{92.74}  & \textbf{94.46} & 59.28 & 66.72 & 68.20\\
			\rowcolor{mygray1}
			C$^2$AM(Ours)$^{\ddag}$ & DenseNet161  & EfficientNet-B7 & 81.76  &  91.11  &  92.88 & \textbf{59.56} & \textbf{67.05} & 68.53 \\
			\bottomrule[1pt]
	\end{tabular}}
	\label{tab:wsol_sota}
	\vspace{-10pt}
\end{table*}

\section{Results and Analysis}
We first present a visual comparison between CAM and the proposed C$^2$AM to validate that more complete and correct foreground regions can be predicted in the class-agnostic activation maps generated by C$^2$AM. Then, we apply C$^2$AM to WSOL and show that CAM can be replaced by the class-agnostic activation maps for more accurate object localization. Besides, we also apply C$^2$AM to WSSS and show that the class-agnostic activation maps can serve as background cues to reduce false-activation of background in CAM and segment more reliable object regions. Details are presented in the following sections.

\begin{table}[!htbp]
	\centering
	\caption{The GT-Known Loc of DDT and our C$^2$AM on the CUB-200-2011 and ImageNet-1K. Init. denotes the parameters initialization. sup-pre.: the parameters initialized from supervised pretraining. moco~\cite{moco} and detco~\cite{detco}: the unsupervised pretraining. $^*$ : the result is obtained with our re-implementation.}
	\vspace{-5pt}
	\resizebox{\linewidth}{!}{
		\begin{tabular}{l c c c c}
			\toprule[1.5pt]
			Method & Bac. & Init. & CUB-200-2011 & ImageNet-1K \\
			\midrule
			DDT & VGG16 & sup-pre. & \textbf{84.55} & 61.41 \\
			\rowcolor{mygray}
			C$^2$AM & VGG16 & sup-pre. & 75.34 & \textbf{63.41} \\
			DDT & ResNet50 & sup-pre. & 72.39 & 59.92 \\
			\rowcolor{mygray}
			C$^2$AM & ResNet50 & sup-pre. & \textbf{89.99} & \textbf{65.89} \\
			\midrule
			DDT$^*$ & ResNet50 & moco & 34.95 & 40.52 \\
			\rowcolor{mygray}
			C$^2$AM & ResNet50 & moco & \textbf{89.90} & \textbf{66.51} \\
			DDT$^*$ & ResNet50 & detco & 35.86 &  41.23 \\
			\rowcolor{mygray}
			C$^2$AM & ResNet50 & detco & \textbf{88.21} & \textbf{65.48} \\
			\bottomrule[1pt]
	\end{tabular}}
	\label{tab:pseudo_bboxes}
	\vspace{-5pt}
\end{table}

\begin{table}[t]
	\centering
	\caption{Evaluation results in terms of MaxBoxAccV2.}
	\vspace{-10pt}
	\resizebox{\linewidth}{!}{
		\begin{tabular}{l c c c c c c}
			\toprule[1.5pt]
			\multirow{2.5}{*}{Methods} &  \multicolumn{3}{c}{ImageNet} & \multicolumn{3}{c}{CUB}\\
			\cmidrule(lr){2-4}
			\cmidrule(lr){5-7}
			& VGG & Inception & ResNet50 & VGG & Inception & ResNet50 \\
			\midrule
			Best WSOL & 60.6 & 63.9 & 63.7 & 66.3 & 58.8 & 66.4 \\
			\rowcolor{mygray}
			C$^2$AM(Ours) & \textbf{66.3} & \textbf{65.8} & \textbf{66.8} & \textbf{81.4} & \textbf{82.4} & \textbf{83.8}  \\
			\bottomrule[1pt]
	\end{tabular}}
	\label{tab:maxboxaccv2}
	\vspace{-10pt}
\end{table}

\begin{table}[!htbp]
	\centering
	\caption{Semantic segmentation performance (IoU(\%)) comparison for 11 categories between the initial CAM and CAM refined by using the proposed C$^2$AM. The initial CAM is generated using PSA, SC-CAM, SEAM, PuzzleCAM, and AdvCAM. respectively. $^{\ddag}$ indicates that unsupervised pretraining (moco) is adopted to initialize the backbone network $h(\cdot)$ of C$^2$AM.}
	\vspace{-5pt}
	\resizebox{\linewidth}{!}{
		\begin{tabular}{l l c c c c c c c c c c}
			\toprule[1.5pt]
			Method & bkg & areo & bike & bird & boat & bottle & bus & car & cat & chair & cow \\
			\midrule
			PSA~\cite{affinitynet} & 78.0 & 41.0 & 27.5 & 42.0 & 34.4 & 44.5 & 63.3 & 53.3 & 43.2 & \textbf{30.9} & 49.3 \\
			\rowcolor{mygray}
			+ C$^2$AM$^{\ddag}$ & 
			\textbf{88.0} & \textbf{75.4} & \textbf{45.1} & \textbf{79.5} & \textbf{52.6} & \textbf{64.8} & \textbf{78.0} & \textbf{73.5} & \textbf{81.4} & 27.0 & \textbf{78.2} \\
			\midrule
			SC-CAM~\cite{sc-cam} & 78.6 & 42.1 & 29.2 & 44.5 & 37.0 & 56.1 & 69.9 & 58.2 & 59.8 & \textbf{27.8} & 52.5 \\
			\rowcolor{mygray}
			+ C$^2$AM$^{\ddag}$ & \textbf{88.0} & \textbf{74.8} & \textbf{46.6} & \textbf{80.5} & \textbf{52.8} & \textbf{68.5} & \textbf{79.0} & \textbf{74.6} & \textbf{84.7} & 26.1 & \textbf{78.3} \\
			\midrule
			SEAM~\cite{seam} & 82.8 & 51.0 & 35.4 & 57.4 & 31.6 & 50.0 & 57.9 & 63.2 & 62.4 & \textbf{27.3} & 63.0 \\	
			\rowcolor{mygray}
			+ C$^2$AM$^{\ddag}$ & \textbf{87.1} & \textbf{71.9} & \textbf{44.2} & \textbf{77.9} & \textbf{45.6} & \textbf{61.9} & \textbf{75.5} & \textbf{70.4} & \textbf{82.5} & 22.5 & \textbf{78.0} \\
			\midrule
			PuzzleCAM~\cite{puzzlecam} & 78.5 & 43.2 & 32.3 & 36.7 & 23.1 & 51.7 & 67.6 & 61.4 & 76.7 & 16.7 & 60.4 \\
			\rowcolor{mygray}
			+ C$^2$AM$^{\ddag}$ & \textbf{87.6} & \textbf{71.5} & \textbf{48.4} & \textbf{78.0} & \textbf{44.6} & \textbf{66.7} & \textbf{77.3} & \textbf{74.3} & \textbf{85.7} & \textbf{21.3} & \textbf{80.2} \\
			\midrule
			AdvCAM~\cite{advcam} & 81.3 & 50.6 & 33.5 & 57.5 & 37.0 & 53.3 & 67.8 & 54.8 & 64.7 & \textbf{35.0} & 68.4 \\
			\rowcolor{mygray}
			+ C$^2$AM$^{\ddag}$ & \textbf{87.5} & \textbf{72.9} & \textbf{46.4} & \textbf{78.9} & \textbf{50.7} & \textbf{60.5} & \textbf{77.8} & \textbf{71.2} & \textbf{84.5} & 26.8 & \textbf{79.5} \\
			\bottomrule[1pt]
	\end{tabular}}
	\label{tab:refined_cam_1}
	\vspace{-12pt}
\end{table}

\subsection{Visual Comparison with CAM}
Figure \ref{fig:visual_compar} presents the visual comparison between CAM and the proposed C$^2$AM on ImageNet-1K and PASCAL VOC2012 datasets. The second and fifth columns present the visualization of CAM. It can be seen that with image-level supervision, CAM can correctly indicate the location of the object but usually focus only on the most discriminative regions of the target object, e.g., the head of bird. Besides, CAM also activates class-related regions, e.g., the railroad, and ignores the complete body of the train. These issues degrade the performance of CAM in WSOL and WSSS. In contrast, the class-agnostic activation maps generated by C$^2$AM effectively mitigate the above issues. As shown in the left of Figure \ref{fig:visual_compar}, C$^2$AM can completely separate the foreground objects out of background regions, in which the whole body of bird is activated. Besides, as shown in the right of Figure \ref{fig:visual_compar}, C$^2$AM successfully discriminate the railroad from the train and thus the complete region of train can be activated. More visual results are provided in supplementary materials. 

\subsection{Results of WSOL}

\textbf{Class-agnostic object bounding boxes.} Table~\ref{tab:pseudo_bboxes} compares the quality of class-agnostic bounding boxes generated by C$^2$AM and DDT~\cite{ddt}. When supervised pretraining on ImageNet-1K are adopted as initialization of $h(\cdot)$, class-agnostic bounding boxes generated by C$^2$AM have a higher GT-known Loc than DDT (except the backbone of VGG16). This indicates that our method has a good capability to generate higher quality class-agnostic bounding boxes than DDT. When considering unsupervised pretraining, e.g., moco~\cite{moco} and detco~\cite{detco}, the GT-known Loc of DDT on two benchmarks has a large drop. By contrast, our C$^2$AM maintains high performance, which is even better than that of DDT using supervised pretraining.

\textbf{Results on CUB-200-2011.} Table~\ref{tab:wsol_sota} compare the performance of C$^2$AM and state-of-the-art methods on CUB-200-2011. Our method achieves Top-1 Loc of 83.28\%, Top-5 Loc of 92.74\%, and GT-known Loc of 94.46\%. When compared with SPA and ORNet, our C$^2$AM surpasses them by a large margin in terms of Top-1 Loc, Top-5 Loc, and GT-known Loc. Compared with PSOL, our C$^2$AM with supervised pretraining surpasses it by 2.39\%, 2.77\%, and 2.68\%, in terms of Top-1 Loc, Top-5 Loc, and GT-known Loc, respectively. When $h(\cdot)$ is initialized by moco, C$^2$AM also achieves better performance than PSOL. We also provide the results of MaxBoxAccV2 metric in Table~\ref{tab:maxboxaccv2}, where the result of Best WSOL is directly taken from \cite{ewsol}. One can observe that our C$^2$AM achieves significantly better performance for different networks and datasets.

\textbf{Results on ImageNet-1K.} Table \ref{tab:wsol_sota} also compares our method with other recently introduced weakly supervised object localization methods on ImageNet-1K validation set. Our method achieves Top-1 Loc of 59.56\%, Top-5 Loc of 67.05\%, and GT-known Loc of 68.53\%, which achieves the state-of-the-art performance. When using the same localization (DenseNet161~\cite{densenet}) and classification backbone (EfficientNet-B7~\cite{efficientnet}), the performance of C$^2$AM on the validation set significantly surpasses PSOL by 1.56\%, 2.03\%, and 2.25\% in terms of Top-1 Loc, Top-5 Loc, and GT-known Loc, respectively.

\begin{table}[t]
	\centering
	\caption{Semantic segmentation performance (mIoU(\%)) comparison among different methods. dCRF: dense CRF~\cite{dcrf}. FSSD: background cues from fully supervised saliency detector~\cite{poolnet}. $^{\ddag}$: moco is adopted to initialize the backbone network $h(\cdot)$ of C$^2$AM.}
	\vspace{-5pt}
	\resizebox{\linewidth}{!}{
		\begin{tabular}{l c c c >{\columncolor{mygray}}l}
			\toprule[1.5pt]
			Method & CAMs & + dCRF & + FSSD & \textbf{+ C$^2$AM$^{\ddag}$} \\
			\midrule
			
			PSA$_\text{ CVPR'18}$~\cite{affinitynet} & 48.0 & - & 62.0 & \textbf{65.5} (+\textcolor{red}{17.5}) \\
			SC-CAM $_\text{ CVPR'20}$~\cite{sc-cam}& 50.9 & 55.3 & 62.1 & \textbf{66.0} (+\textcolor{red}{15.1}) \\
			SEAM$_\text{ CVPR'20}$~\cite{seam} & 55.4 & 56.8 & 61.5 & \textbf{63.9} (+\textcolor{red}{8.5}) \\
			PuzzleCAM$_\text{ ICIP'21}$~\cite{puzzlecam} & 51.5 & - & 62.1 & \textbf{65.5} (+\textcolor{red}{14.0}) \\
			AdvCAM$_\text{ CVPR'21}$~\cite{advcam} & 55.6 & 62.1 & 62.3 & \textbf{65.4} (+\textcolor{red}{9.8}) \\
			\bottomrule[1pt]
	\end{tabular}}
	\label{tab:refined_cam_2}
	\vspace{-12pt}
\end{table}

\subsection{Results of WSSS}
\textbf{Results on PASCAL VOC2012.} As this section is designed to demonstrate the effectiveness of C$^2$AM to refine initial CAM and the refined CAM can be further used in other stages of WSSS, the IoU and mIoU are reported on the training set of PASCAL VOC2012.  Table \ref{tab:refined_cam_1} presents improvement of each category for existing CAM-based methods by using C$^2$AM. It can be seen that the segmentation performances of many CAM-based methods for most categories, including PSA~\cite{affinitynet}, SC-CAM~\cite{sc-cam}, SEAM~\cite{seam}, PuzzleCAM~\cite{puzzlecam}, and AdvCAM~\cite{advcam},  are largely improved by using C$^2$AM. For example, there is a mean of 14.6\%, 31.3\%, 20.1\% IoU improvements of these five methods for bike, bird, and cow, respectively.

To compare our C$^2$AM with other CAM refinement approaches, Table \ref{tab:refined_cam_2} presents the segmentation performance of initial CAMs generated by literature works like PSA, SC-CAM, and PuzzleCAM etc., and the improvements after they are refined by dense CRF (+dCRF)~\cite{dcrf}, and background cues generated by a fully supervised saliency detector (+FSSD)~\cite{poolnet}, and our C$^2$AM (+C$^2$AM). The results of dense CRF are reported in their papers and the refinement process
presented in Section~\ref{sec:wsss} is used for both +FSSD and +C$^2$AM. It can be clearly seen that the quality of initial CAM has been greatly improved by using our C$^2$AM. In particular, the mIoU of initial CAM generated by PSA~\cite{affinitynet} and SC-CAM~\cite{sc-cam} have been improved by 17.5\%
and 15.1\%, respectively. In particular,  the increase of mIoU caused by background cues generated from C$^2$AM is even higher than that generated by the fully supervised saliency detector.

We also provide a visualization of CAM refinement in Figure \ref{fig:cam_refinement}. It can be observed in the first column that, a lot of object regions, e.g., the body of the white cow and the feet of the brown horse, are not activated in the initial CAM. Besides, backgrounds, e.g., the grass and ground, are usually falsely activated. To address these issues, we use the class-agnostic activation maps to extract background cues (as shown in the second column) to efficiently reduce the false activation of background and predict more reliable object regions. The results shown in the third column highly overlaps with the ground truth (in the last column).

\begin{figure}[t]
	\begin{center}
		\includegraphics[width=\linewidth]{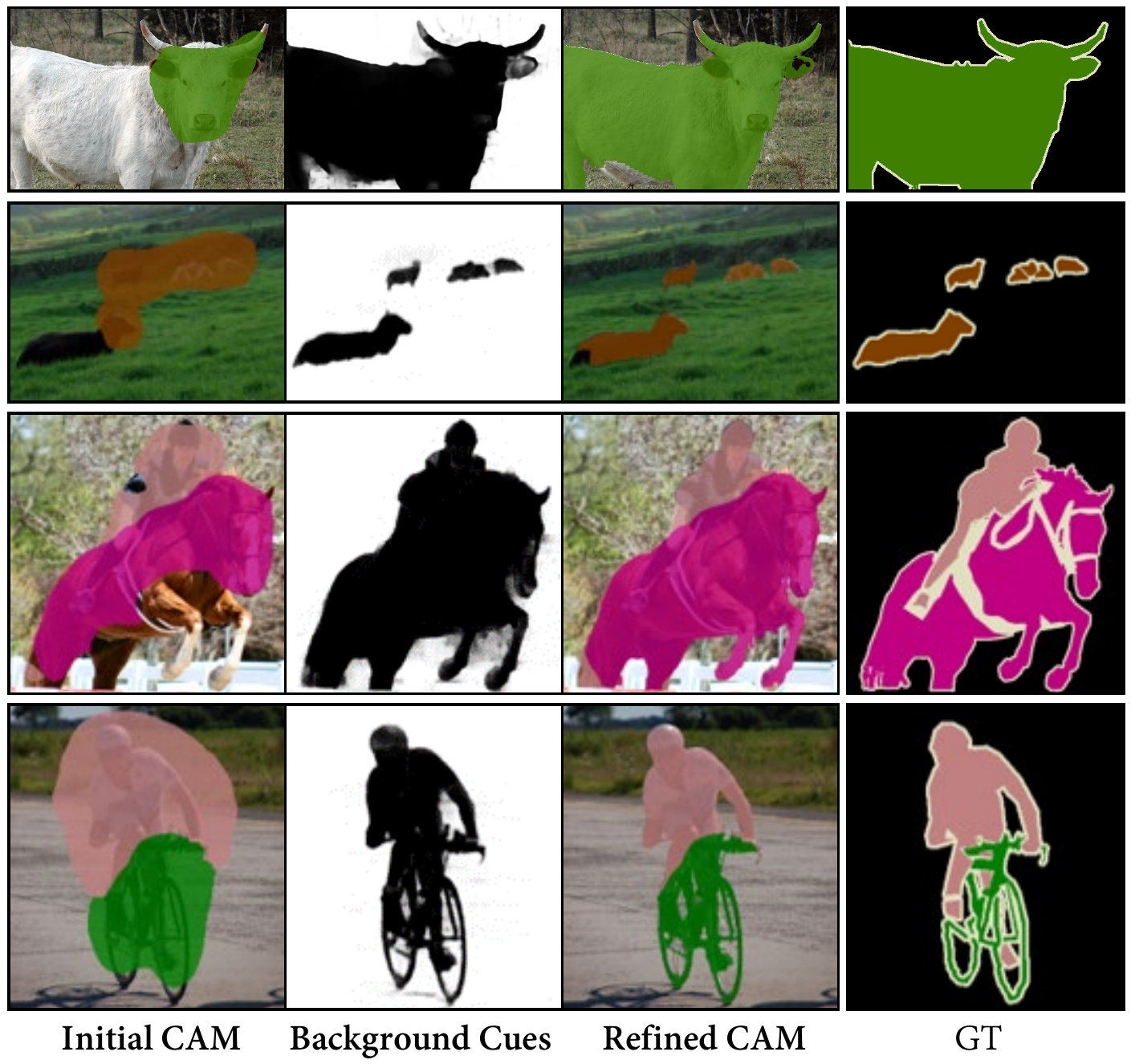}
	\end{center}
	\vspace{-18pt}
	
	\caption{Illustration of CAM refinement using background cues extracted from C$^2$AM. First column: initial CAM. Second column: background cues extracted from C$^2$AM (255: background, 0: foreground). Third column: CAM refined using background cues. Last column: ground-truth masks. Best viewed in color.}
	\label{fig:cam_refinement}
	\vspace{-8pt}
\end{figure}

\begin{table}[t]
	\centering
	\caption{Sensitivity analysis of parameters initialization of 3$\times$3 convolution in activation head. k\_normal and k\_uniform are short for kaiming\_normal and kaiming\_uniform initialization in PyTorch~\cite{paszke2019pytorch}. Results are reported on ImageNet-1K validation set.}
	\vspace{-5pt}
	\resizebox{\linewidth}{!}{
		\begin{tabular}{c c c c c}
			\toprule[1.5pt]
			Metric & k\_normal & k\_uniform & normal & uniform \\
			\midrule
			\rowcolor{mygray}
			GT-known Loc & 66.51 & 66.44 & 66.82 & 66.67 \\
			\bottomrule[1pt]
	\end{tabular}}
	\label{tab:param_init}
	\vspace{-12pt}
\end{table}
\subsection{Sensitivity Analysis}
We conduct experiments on ImageNet-1K to examine the hyper-parameter sensitivity of C$^2$AM. The GT-known Loc is adopted as a performance evaluation.

\textbf{Initialization of Activation Head.} A 3$\times$3 convolution with a batch normalization layer is adopted as the activation head $\varphi(\cdot)$. The backbone network $h(\cdot)$ can be initialized with supervised or unsupervised pretraining (e.g., moco, detco). How about the activation head? Does the initialization of activation head affect the performance of C$^2$AM? We perform experiments with four different parameters initialization of activation head, i.e., kaiming\_normal, kaiming\_uniform, normal, and uniform initialization. The weights and bias in batch normalization layer are initialized with 1 and 0. Table \ref{tab:param_init} compares the performance of these four initialization choices. We can see that the performances are quite stable among these initialization methods, which shows that the activation head of C$^2$AM is not sensitive to the parameters initialization. 

\textbf{Hyper-parameter $\alpha$.} Figure \ref{fig:alpha_sens} presents the GT-known Loc of C$^2$AM with different setting of $\alpha$ in Eq.~\ref{eq:rank_weighting}. The results are reported on ImageNet-1K validation set. When $\alpha$ equals 0, every $w_{i,j}$ is equal to 1, and every positive pair including those foreground with different appearance and background with significant different color/texture are also pulled together. This violates the assumption that only representations with similar appearance, color or texture can be pulled close. Thus, the GT-known Loc is the lowest when $\alpha$ is set to 0. When $\alpha$ is larger than 0, different positive pairs are assigned with different weights based on the feature similarity. As seen, the Gt-Known Loc has been largely improved when $\alpha$ is larger than 0, demonstrating that the proposed rank weighting works well to automatically reduce the influence of those dissimilar positive pairs. When $\alpha$ varies from 0.1 to 0.9, the number of positive pairs involved for contrastive learning is gradually decreasing. The stable performances show that our approach is not sensitive to the number of positive pairs (i.e., the setting of $\alpha$).

\begin{figure}[t]
	\begin{center}
		\includegraphics[width=0.85\linewidth]{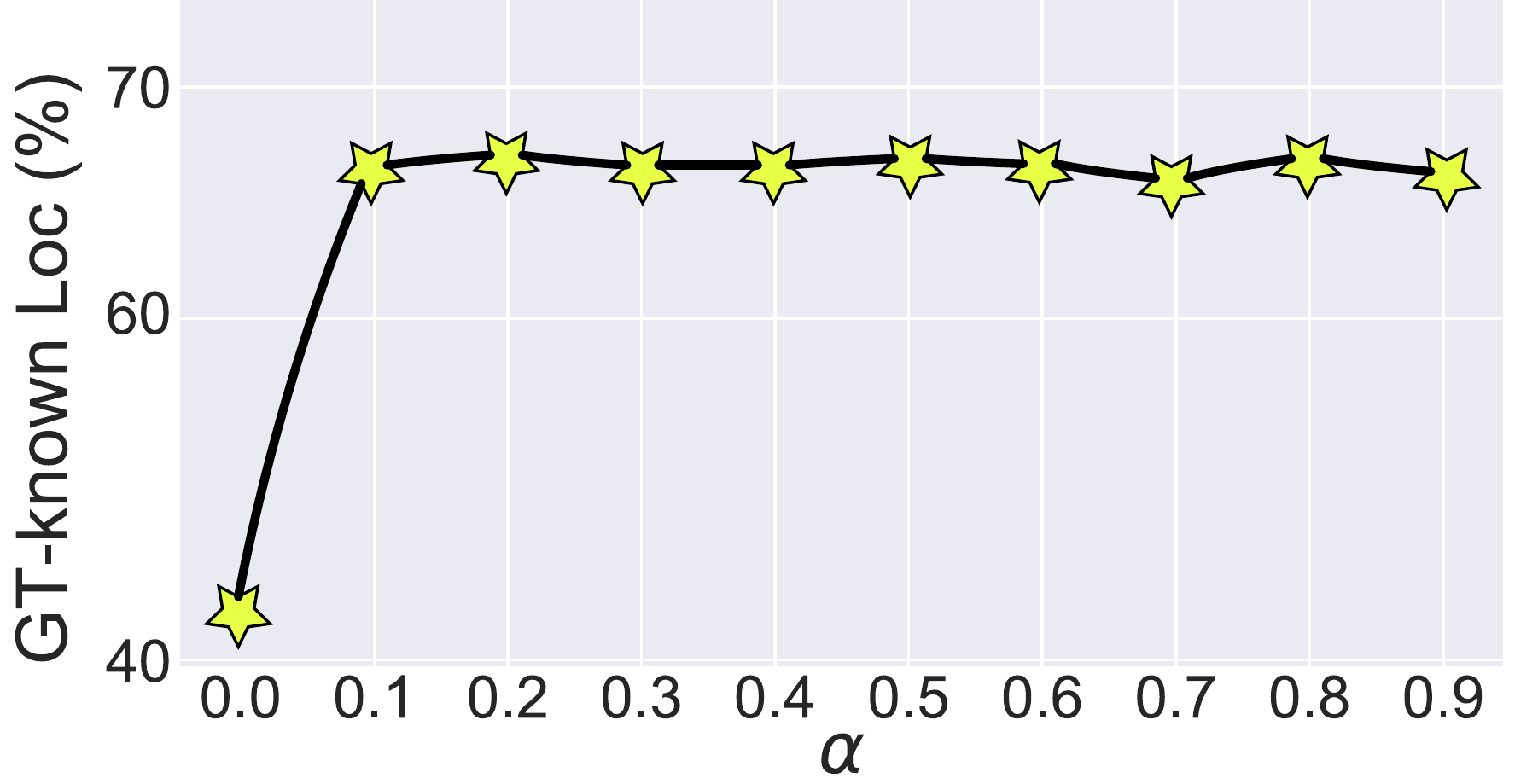}
	\end{center}
	\vspace{-18pt}
	\caption{Sensitivity analysis of hyper-parameter $\alpha$ in Eq.~\ref{eq:rank_weighting}. The GT-known Loc is reported on ImageNet-1K validation set.}
	\label{fig:alpha_sens}
	\vspace{-14pt}
\end{figure}

\section{Conclusions and Discussions}
We propose cross-image foreground-background contrast for class-agnostic activation maps generation using unlabeled image data. Class-agnostic activation map determines more reliable foreground regions, which can be used to replace or refine CAM for better WSOL and WSSS performance. Extensive experiments show that both WSOL and WSSS can benefit from our approach. We provide an alternative solution for performance improvements of weakly supervised learning.

As the proposed C$^2$AM works well in the unlabeled training data, we believe that it can be further used to efficiently detect the foreground regions without manual annotations in a lot of vision tasks, e.g., saliency detection and skin lesion segmentation, etc.

{\small
\bibliographystyle{ieee_fullname}
\bibliography{egbib}
}

\end{document}